\documentclass{article}

\usepackage{arxiv}

\usepackage[utf8]{inputenc} 
\usepackage[T1]{fontenc}    
\usepackage{hyperref}       
\usepackage{url}            
\usepackage{booktabs}       
\usepackage{amsfonts}       
\usepackage{nicefrac}       
\usepackage{microtype}      
\usepackage{lipsum}		
\usepackage{graphicx}
\usepackage{doi}

\usepackage{algorithm}
\usepackage{algpseudocode}
\usepackage{xcolor}
\usepackage{amssymb,amsmath}
\usepackage[small]{caption}
\usepackage{multirow}
\usepackage{subfigure}
\usepackage{bm}
\usepackage{bbm}
\usepackage{comment}
\usepackage{wrapfig}
\usepackage{tikz}
\usepackage{pgfplots}
\usepackage{appendix}

\title{A User-Guided Bayesian Framework for Ensemble Feature Selection in Life Science Applications (UBayFS)}

\author{ \href{https://orcid.org/0000-0002-6919-3483}{\includegraphics[scale=0.06]{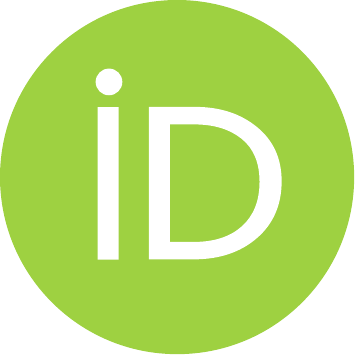}\hspace{1mm}Anna Jenul }\thanks{both authors contributed equally to the work}\\
  Department of Data Science\\
  Norwegian University of Life Sciences\\
  \texttt{anna.jenul@nmbu.no} \\
   \And
 \href{https://orcid.org/0000-0003-1327-4855}{\includegraphics[scale=0.06]{orcid.pdf}\hspace{1mm}Stefan Schrunner\footnotemark[1]} \\
  Department of Data Science\\
  Norwegian University of Life Sciences\\
  \texttt{stefan.schrunner@nmbu.no} \\
  \And
 \href{https://orcid.org/0000-0001-9365-4916}{\includegraphics[scale=0.06]{orcid.pdf}\hspace{1mm}Jürgen Pilz}\\
  Department of Statistics\\
  Norwegian University of Life Sciences\\
  \texttt{juergen.pilz@aau.at} \\
  \And
 \href{https://orcid.org/0000-0003-1595-9962}{\includegraphics[scale=0.06]{orcid.pdf}\hspace{1mm}Oliver Tomic}\\
  Department of Data Science\\
  Norwegian University of Life Sciences\\
  \texttt{oliver.tomic@nmbu.no} \\
}

\renewcommand{\shorttitle}{User-Guided Bayesian Framework for Ensemble Feature Selection (UBayFS)}

\hypersetup{
pdftitle={UBayFS preprint},
pdfauthor={Anna Jenul},
pdfkeywords={Ensemble feature selection, Bayesian model, Dirichlet-multinomial, user constraints},
}

\begin{document}
\maketitle

\begin{abstract}
Feature selection represents a measure to reduce the complexity of high-dimensional datasets and gain insights into the systematic variation in the data. This aspect is of specific importance in domains that rely on model interpretability, such as life sciences. We propose UBayFS, an ensemble feature selection technique embedded in a Bayesian statistical framework. Our approach considers two sources of information: data and domain knowledge. We build a meta-model from an ensemble of elementary feature selectors and aggregate this information in a multinomial likelihood. The user guides UBayFS by weighting features and penalizing specific feature blocks or combinations, implemented via a Dirichlet-type prior distribution and a regularization term. In a quantitative evaluation, we demonstrate that our framework (a) allows for a balanced trade-off between user knowledge and data observations, and (b) achieves competitive performance with state-of-the-art methods.
\end{abstract}

\keywords{Ensemble feature selection \and Bayesian model \and Dirichlet-multinomial \and user constraints}

\section{Introduction}\label{sec1}

Feature selection pursues two major goals: to improve generalizability and performance of predictive algorithms like classification, regression, or clustering models and to improve data understanding and interpretability. Both aspects are of significant interest in fields like healthcare, where major decisions may be based on data analysis. Here, two sources of information are available: large-scale collections of data from multiple sources and profound knowledge from domain experts. Previous works tend to handle these sources as opposites, see \cite{cheng:expertvsFS}, or neglect expert knowledge completely, see \cite{pozzoli:medicalFSwithDomainKnowledge}. However, a combination of both can be valuable to compensate for underdetermined problem setups from high-dimensional datasets. Moreover, meta-information on the feature set may leverage interpretability. Works such as \cite{liu:Constraints} consider constraints between samples but neglect constraints between features. The extension of L1 regularization to the so-called \textit{Group Lasso} \cite{yuan:groupLasso} and its variants \cite{ida:fSparseGlasso} account for block structure but cannot handle more complex constraint types. There is a lack of sophisticated probabilistic frameworks that tackle this issue and deliver transparent results.

Apart from measuring the influence on model performance, properties like stability and reproducibility of the feature selector are essential to ensure that the user can trust the predictive model. Even though variants to achieve reproducibility are available for certain model types, such as deep neural networks \cite{lu:reproducibleFS}, a model-independent approach to stabilizing the feature selection process is to deploy ensembles of elementary feature selectors. Recent research pursued this idea by \cite{jenul:rent} utilizing regularized linear or generalized linear models and involving measures for stability in addition to predictive performance metrics. \cite{seijo:ensembleSurvey} conclude that meta-models composed of elementary feature selectors improve the performance and robustness of the selected feature set in many cases. However, to the best of our knowledge, probabilistic approaches that exploit both --- a sound statistical framework and individual model benefits of using an ensemble elementary feature selectors --- are not yet available.

A prominent framework with the capability to combine data and expert knowledge is Bayesian statistics, which has been applied for feature selection in linear models, see \cite{hara:review_bayes}. Intentions behind the usage of Bayesian methodology vary significantly between authors and do not necessarily involve expert knowledge. Examples include \cite{dalton:medicalBayesianFS}, who investigate sparsity priors and \cite{goldstein:medicalFS}, who suggest a Bayesian framework to quantify the level of uncertainty in the underlying feature selection model. Other Bayesian approaches for feature selection include \cite{lyle:bayesModelSelection}, and \cite{saon:BayesErrorFS}, but these works do not investigate the usage of expert knowledge as prior. Although the availability of expert knowledge plays a role in life sciences, none of these approaches strongly emphasize domain knowledge about features, nor do they involve specific prior constraints defined by the user.

In this work, we propose a novel Bayesian approach to feature selection that incorporates expert knowledge and maintains large model generality. We aim to fill the gap between data-driven feature selection on one side and purely expert focused feature selection on the other side. Our presented probabilistic approach, UBayFS, combines a generic ensemble feature selection framework with the exploitation of domain knowledge, such that it supports interpretability and improves the stability of the results. For this purpose, feature importance votes from independent elementary feature selectors are merged with constraints and feature weights specified by the expert. Constraints may be of a general type, such as a maximum number of features or blocks of features to be selected. Both inputs, likelihood and prior, are aggregated in a sound statistical framework, producing a posterior probability distribution over all possible feature sets. We use a Genetic Algorithm for discrete optimization to efficiently optimize the posterior feature set in high-dimensional datasets. In an extensive experiment section, we analyze UBayFS in a case study covering a variety of potential model constraints and parameter settings. Results on open-source datasets are benchmarked against state-of-the-art feature selectors concerning predictive performance and stability, underlining the potential of UBayFS.

\paragraph{Notations}
We will denote vectors by bold, uncapitalized, and matrices by bold, capitalized letters. Non-bold, uncapitalized letters indicate scalars or functions, and non-bold, capitalized letters indicate sets or constants. $\Vert.\Vert_1$ denotes the $L1$-norm. $[N]$ is an abbreviation of the set of indices ${1,\dots,N}$. The $N$-dimensional vector of ones will be written as $\mathbbm{1}_N$. Furthermore, we refer to sets of features by their feature indices, such as $S\subseteq [N]$, or by a binary membership vector $\bm{\delta}^S\in\{0,1\}^N$ with components $(\bm{\delta}^S)_n = \left\{\begin{array}{ll} 1 & \text{if}~n\in S, \\ 0 & \text{otherwise.}\end{array}\right.$

\section{User-Guided Ensemble Feature Selector}
\label{sec:method}

Given a finite set of $N$ features, the goal of UBayFS is to find an optimal subset of feature indices $S^{\star}\subset [N]$, or equally $\bm{\delta}^{\star}\in\{0,1\}^N$. We assume that information is available from
\begin{enumerate}
    \item training data to collect evidence by conventional data-driven feature selectors---we denote this as \textit{information from data} $\bm{\Delta}$, 
    \item the user's domain knowledge encoded as subjective beliefs $\bm{\alpha}\in\mathbb{R}^N$ about the importance of features, where $\alpha_n>0$ for all $n\in[N]$, and 
    \item side constraints $\bm{A}\bm{\delta}\leq \bm{b}$ to ensure that the obtained feature set conforms with practical requirements and restrictions.
\end{enumerate}
The proposed probabilistic model, UBayFS, builds on the definition of a loss function $L$, which evaluates the quality of selecting a feature set $\bm{\delta}\in\{0,1\}^N$ in the presence of a vector of feature importances $\bm{\theta}\in\Theta$, where $\Theta = \{\bm{\theta}\in[0,1]^N: \Vert \bm{\theta}\Vert_1=1\}$. The parameter vector $\bm{\theta}$ is assumed to be probabilistic and not directly observable, such that evidence about $\bm{\theta}$ is collected from data and prior weights. In specific, $L:\{0,1\}^N\times \Theta\rightarrow \mathbb{R}^+$ links the unknown feature importances to the decision to select a feature set $\bm{\delta}$. We define $L$ in the following way:
\begin{equation} L(\bm{\delta}, \bm{\theta}) = (\bm{1}_N-\bm{\delta})^T\bm{\theta} + \lambda \cdot \left( 1- \kappa(\bm{\delta};\bm{A},\bm{b},\bm{\rho})\right),
\end{equation}
where $\bm{1}_N$ denotes the $N$-dimensional vector of ones, $\kappa$ is a function accounting for violations of the side constraints, and $\lambda > 0$ indicates the overall power of the constraints (the purpose of $\bm{\rho}$ will be discussed at a later point along with the formulation of the constraint function $\kappa$). Thus, $L$ accumulates the importances of all non-selected features (residual information) and penalizes the violation of side constraints $\bm{A}\bm{\delta}\leq \bm{b}$ via a regularization term.

In terms of statistical decision theory, decisions should minimize the risk $r(\bm{\delta})$, which is given as the expected loss function over all possible states of nature $\bm{\theta}$:
\begin{align}
 r(\bm{\delta}) &= \mathbb{E}_{\bm{\theta}}\left[ L(\bm{\delta}, \bm{\theta}) \right] = (\bm{1}_N - \bm{\delta})^T\mathbb{E}_{\bm{\theta}}\left[ \bm{\theta} \right] + \lambda\cdot \left(1-\kappa(\bm{\delta};\bm{A},\bm{b},\bm{\rho})\right) \longrightarrow \underset{\bm{\delta}\in\{0,1\}^N}{\min} \label{eq:loss} \\
 &\Leftrightarrow \bm{1}_N^T\mathbb{E}_{\bm{\theta}}\left[ \bm{\theta} \right] - \bm{\delta}^T\mathbb{E}_{\bm{\theta}}\left[ \bm{\theta} \right] + \lambda - \lambda \kappa(\bm{\delta};\bm{A},\bm{b},\bm{\rho})\longrightarrow \underset{\bm{\delta}\in\{0,1\}^N}{\min} \\
 &\Leftrightarrow \bm{\delta}^T\mathbb{E}_{\bm{\theta}}\left[ \bm{\theta} \right] + \lambda \kappa(\bm{\delta};\bm{A},\bm{b},\bm{\rho})\longrightarrow \underset{\bm{\delta}\in\{0,1\}^N}{\max}
\end{align}
To determine $\mathbb{E}_{\bm{\theta}}[\bm{\theta}]$ accordingly, UBayFS evaluates data from elementary feature selectors trained on subsets of the dataset, summarized as $\bm{\Delta}$, as well as prior feature importance scores $\bm{\alpha}$. Thus, the posterior probability distribution over the unknown feature importance parameter $\bm{\theta}$ given the independent data sources $\bm{\Delta}$ and $\bm{\alpha}$, $p(\bm{\theta} \vert \bm{\Delta},\bm{\alpha})$, is decomposed using Bayes' theorem into
\begin{equation} 
    p(\bm{\theta} \vert \bm{\Delta},\bm{\alpha}) \propto p(\bm{\Delta} \vert \bm{\theta}) \cdot p(\bm{\theta} \vert \bm{\alpha})
\end{equation}
where $p(\bm{\Delta} \vert \bm{\theta})$ describes the model likelihood (evidence from elementary feature selector models) and $p(\bm{\theta} \vert \bm{\alpha})$ describes the density of a prior distribution (user knowledge). The core part of UBayFS is to derive parametrizations for likelihood and prior distribution from our model inputs. Due to the convenient representation of the loss function, Eq. \ref{eq:loss}, it suffices to determine the expected value of the posterior distribution of $\bm{\theta}$. The optimal feature set is then given by
\begin{equation}
    \bm{\delta}^\star = \underset{\bm{\delta}\in \{0,1\}^N}{\text{arg}~\min}~r(\bm{\delta}),
\end{equation}
which can be solved numerically via discrete optimization.

\subsection{Ensemble feature selection as likelihood}

To collect information about feature importances from the given dataset, we train an ensemble of $M$ elementary feature selectors of the same model type on distinct training subsets. The selection of a feature index set $\bm{\delta}^{(m)}$ comprising a number of $l = \Vert \bm{\delta}\Vert_1$ features in each elementary model $m$ out of a total of $M$ models can be interpreted as a result of drawing $l$ balls from an urn, where each ball has a distinct color representing one feature $n\in [N]$. Over all elementary models, $\bm{\Delta}$ collects the counts of each feature being selected, resulting in a count vector in \begin{equation}
    \bm{\Delta} = \sum\limits_{m=1}^{M}\bm{\delta}^{(m)}\in\{0,\dots,M\}^N.
\end{equation}
Each elementary feature selector delivers a proposal for an optimal feature set. Thus, we let the frequency of drawing a feature throughout $\bm{\delta}^{(1)},\dots,\bm{\delta}^{(M)}$ represent its \textit{importance} by defining the latent importance parameter vector $\bm{\theta} \in [0,1]^N$, $\Vert \bm{\theta}\Vert_1 = 1$, as the success probabilities of sampling each feature in an individual urn draw. In a statistical sense, we interpret the result from each elementary feature selector as realization from a multinomial distribution with parameters $\bm{\theta}$ and $l$.\footnote{The exact way to describe this procedure is a multivariate hypergeometric distribution, since each feature occurs at most once in a set, but an approximation using the multinomial distribution facilitates computation.} This multinomial setup delivers the likelihood $p(\bm{\Delta} \vert \bm{\theta})$ as joint probability density
\begin{align}
    p(\bm{\Delta} \vert \bm{\theta}) = \prod\limits_{m = 1}^{M} f_{\text{mult}}(\bm{\delta}^{(m)};\bm{\theta},l),
\end{align}
where $f_{\text{mult}}(\bm{\delta}^{(m)};\bm{\theta},l)$ denotes the density of a multinomial distribution with success probabilities $\bm{\theta}$ and a number of $l$ urn draws. Relevant notations are summarized in Tab. \ref{tab:glossary1}.

\begin{table}[ht]
    \centering
    \caption{Notations for likelihood parameters.}
    \label{tab:glossary1}
    \begin{tabular}{l l}
    \toprule
    \multicolumn{2}{c}{input \& elementary models} \\
    \midrule
    $n \in [N]$ & feature indices \\
    $m \in [M]$ & elementary models \\
    $\bm{\delta}\in \{0,1\}^N$ & feature index set \\
    $\bm{\theta}\in \Theta \subset [0,1]^N$ & feature importances \\
    $\bm{\Delta}\in \{0,\dots,M\}^N$ & feature counts \\
    \bottomrule
    \end{tabular}
\end{table}

\subsection{Expert knowledge as prior weights}
\label{sec:prior}
To constitute the prior distribution, UBayFS uses expert knowledge as a-priori weights of features. Since the domain of the distribution of feature importances $\bm{\theta}$ is defined to be a simplex $\bm{\theta}\in \Theta \subset [0,1]^N, \Vert \bm{\theta}\Vert_1 = 1$, the Dirichlet distribution is a natural choice as prior distribution, which is widely used in data science problems, such as \cite{nakajima:DirichletFS}. Thus, we initially assume that a-priori
\begin{equation}
    p(\bm{\theta}) = f_{\text{Dir}}(\bm{\theta};\bm{\alpha}),
\end{equation}
where $f_{\text{Dir}}(\bm{\theta};\bm{\alpha})$ denotes the density of the Dirichlet distribution with positive $\bm{\alpha} = (\alpha_1,\dots,\alpha_N)$. Since the Dirichlet distribution is a conjugate prior of the multinomial distribution, the posterior distribution results in a Dirichlet type, again, see \cite{degroot2005optimal}. Thus, it holds for the posterior density that
\begin{equation}
    p(\bm{\theta} \vert \bm{\Delta})\propto f_{\text{Dir}}(\bm{\theta}; \bm{\alpha}^{\circ}),
\end{equation}
where the parameter update is obtained in closed form by
\begin{equation}
    \bm{\alpha}^{\circ} = \bm{\alpha} + \bm{\Delta}.
\end{equation}
In case of integer-valued prior weights $\bm{\alpha}$, they may be interpreted as pseudo-counts in the context of modelling success probabilities in an urn model---comparable to the information gained if the corresponding counts were observed in a multinomial data sample. In UBayFS, we obtain $\bm{\alpha}$ as feature weights provided by the user. If no user knowledge is available, the least informative choice is to specify uniform counts with a small positive value, such as $\bm{\alpha}_{\text{unif}}=0.01\cdot \mathbbm{1}_N$.

\paragraph{Generalized Dirichlet model}
Even though the presented Dirichlet-multinomial model is a popular choice due to its favorable statistical properties, it implicitly assumes that classes are mutually independent. However, high-dimensional datasets frequently involve complex correlation structures between the features. To account for this aspect, we generalize the setup by replacing the Dirichlet prior distribution with some generalized Dirichlet distribution. The highest level of generalization is achieved by \cite{hankin}, who introduce the hyperdirichlet distribution, which may take arbitrary covariance structures into account. The hyperdirichlet distribution maintains the conjugate prior property with respect to the multinomial likelihood, and thus, inference is tractable; however, the analytical expression of the expected value involves the intractable normalization constant and, as a result, requires numerical means such as Monte-Carlo Markov Chain (MCMC) methods, which may face computational challenges due to the high dimensionality of the problem.

A compromise between the complexity of the problem and the flexibility of the covariance structure is given by an earlier version of the generalized Dirichlet distribution by \cite{wong98}, which is a special case of the hyperdirichlet setup, but more general than the standard Dirichlet distribution. In addition to the properties of the hyperdirichlet distribution, the expected value of the generalized Dirichlet distribution can be directly evaluated from the distribution parameters. Section \ref{sec:experiments} provides an experimental evaluation of the proposed variants to account for covariance structures in the UBayFS model.\footnote{Details on the generalized prior distributions are provided in Appendix \ref{secA1}.}

\subsection{Side constraints as regularization}
\label{sec:side_constraints}
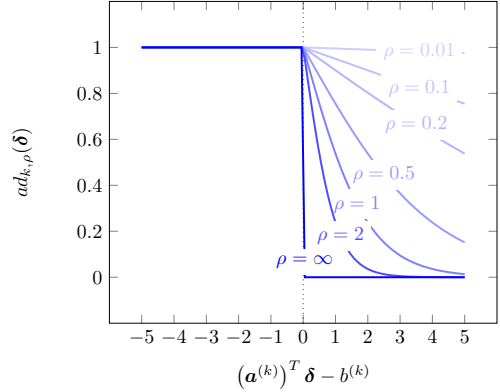
\begin{figure}[ht]
    \centering
    \resizebox{0.4\columnwidth}{!}{
    \begin{tikzpicture}
    \begin{axis}[samples=100,
                 xmin=-6,xmax=6,ymin=-0.2,ymax=1.2,
                 xlabel={$\left(\bm{a}^{(k)}\right)^T\bm{\delta} - b^{(k)}$},  ylabel={$ad_{k,\rho}(\bm{\delta})$},
                 xlabel style={below},ylabel style={above},
                 xtick={-5,-4,...,5},ytick={0,0.2,...,1}]
        \addplot [mark=none, dotted, black] coordinates {(0, -0.2) (0, 1.2)};
        \addplot [mark=none,domain=-5:5, color = blue!20, line width=1] {min(1, 2 * exp(-0.01 * \x) / (1 + exp(-0.01 * \x)))} node[pos=0.86,fill=white] {$\rho = 0.01$};
        \addplot [mark=none,domain=-5:5, color = blue!25, line width=1] {min(1, 2 * exp(-0.1 * \x) / (1 + exp(-0.1 * \x)))} node[pos=0.86,fill=white] {$\rho = 0.1$};
        \addplot [mark=none,domain=-5:5, color = blue!30, line width=1] {min(1, 2 * exp(-0.2 * \x) / (1 + exp(-0.2 * \x)))} node[pos=0.85,fill=white] {$\rho = 0.2$};
        \addplot [mark=none,domain=-5:5, color = blue!40, line width=1] {min(1, 2 * exp(-0.5 * \x) / (1 + exp(-0.5 * \x)))} node[pos=0.75,fill=white] {$\rho = 0.5$};
        \addplot [mark=none,domain=-5:5, color = blue!50, line width=1] {min(1, 2 * exp(-1 * \x) / (1 + exp(-1 * \x)))} node[pos=0.67,fill=white] {$\rho = 1$};
        \addplot [mark=none,domain=-5:5, color = blue!70, line width=1] {min(1, 2 * exp(-2 * \x) / (1 + exp(-2 * \x)))} node[pos=0.625,fill=white] {$\rho = 2$};
        \addplot [mark=none,domain=-5:5, color = blue, line width=1] {(\x < 0)} node[pos=0.54,fill=white] {$\rho = \infty$};
    \end{axis}
    \end{tikzpicture}}
    \caption{The effect of $\rho$ on $ad_{k,\rho}$ for soft constraints.}
    \label{fig:rho}
\end{figure}

Practical setups may require that a selected feature set fulfills certain consistency requirements. These may involve a maximum number of selected features, a low mutual correlation between features, or a block-wise selection of features. UBayFS enables the feature selection model to account for such requirements via a system of $K$ inequalities restricting the feature set $\bm{\delta}$, given as $\bm{A}\bm{\delta}-\bm{b}\leq 0$, where $\bm{A}\in \mathbb{R}^{K\times N}$ and $\bm{b}\in \mathbb{R}^{K}$. Each single constraint $k\in [K]$ can be evaluated via an admissibility function $ad_k(.)$, such that
\begin{equation}
    ad_k(\bm{\delta}) = \left\{\begin{array}{l l}
    1 & \text{if}~ \left(\bm{a}^{(k)}\right)^T\bm{\delta} - b^{(k)} \leq 0 \\
    0 & \text{otherwise},\end{array}\right.
\end{equation}
where $\bm{a}^{(k)}$ is the $k$-th row vector of $\bm{A}$ and $b^{(k)}$ the $k$-th element of $\bm{b}$. UBayFS generalizes the setup by relaxing the constraints: in case that a feature set $\bm{\delta}$ violates a constraint, it shall be assigned a higher penalty rather than being excluded completely. This effect is achieved by replacing $ad_k(.)$ with a relaxed admissibility function $ad_{k,\rho}(.)$ based on a logistic function with relaxation parameter $\rho \in\mathbb{R}^{+}\cup\{\infty\}$: 
\begin{equation}
    ad_{k,\rho}(\bm{\delta}) = \left\{
    \begin{array}{l l}
    1 & \text{if}~\left(\bm{a}^{(k)}\right)^T\bm{\delta}\leq b^{(k)}\\
    0 & \text{if}~ \left(\bm{a}^{(k)}\right)^T\bm{\delta}> b^{(k)} \land \rho =\infty\\
    \frac{2\xi_{k,\rho}}{1 + \xi_{k,\rho}} & \text{otherwise},
    \end{array}
    \right.
\end{equation}
with $\xi_{k,\rho} = \exp\left(-\rho \left(\left(\bm{a}^{(k)}\right)^T \bm{\delta} - b^{(k)}\right)\right)$. Fig. \ref{fig:rho} illustrates that a large parameter $\rho\longrightarrow \infty$ lets the admissibility converge towards the associated hard constraint. A low $\rho$ changes the shape of the penalization to an almost constant function in a local neighborhood around the decision boundary, such that only a minor difference is made between feature sets that fulfill and those that violate a constraint.\footnote{for a proof see Appendix \ref{secA1}}

Finally, the joint admissibility function $\kappa(.)$ aggregates information from all constraints
\begin{equation} 
    \kappa(\bm{\delta}) 
    = \prod\limits_{k=1}^{K} ad_{k,\rho}(\bm{\delta}).
\end{equation}
Note that different relaxation parameters can be specified to prioritize the constraints among each other, hence $\kappa$ involves a parameter vector $\bm{\rho}=(\rho_1,\dots,\rho_K)$. Relevant notations for prior parameters are summarized in Tab. \ref{tab:glossary2}.

\begin{table}[t]
    \caption{Notations used for prior parameters.}
    \label{tab:glossary2}
    \centering
    \begin{tabular}{l l}
    \toprule
    \multicolumn{2}{c}{prior parameters} \\
    \midrule
    $\bm{\alpha}, \bm{\alpha}^\circ \in \mathbb{R}^N$ & prior/posterior weights \\
    $k \in [K]$ & constraint index\\
    $\bm{A}\in \mathbb{R}^{K\times N}$, $\bm{b}\in \mathbb{R}^{K}$ & inequality system \\
    $\bm{\rho} \in \mathbb{R}^{K}$ & relaxation parameters\\
    $\kappa(.):\{0,1\}^N\rightarrow[0,1]$ & joint admissibility\\
    \bottomrule
    \end{tabular}
\end{table}

\paragraph{Feature decorrelation constraints}
Commonly, feature sets with low mutual correlations are preferred since they tend to contain less redundant information. A special case of prior constraints can be defined to enforce that such feature sets are selected. We will refer to such constraints as decorrelation constraints. Decorrelation constraints are pairwise cannot-link constraints between features with high pairwise correlation coefficients---this is achieved by appending a vector $\bm{a}$ with elements 
\begin{equation}
    a_{n} = \left\{\begin{array}{ll} 1 & \text{if }n \in\{i,j\} \\ 0 & \text{else,}\end{array}\right.
\end{equation} 
and an element $b = 1$ to the constraint system. We select the shape parameter $\rho_{i,j}$ for the constraint between features $i$ and $j$ by the odds ratio of the absolute correlation coefficient $\tau_{i,j}$,
\begin{equation}
    \rho_{i,j} = \left\{\begin{array}{ll}\frac{\vert \tau_{i,j}\vert }{1-\vert \tau_{i,j}\vert } & \text{if }\vert \tau_{i,j}\vert > \tau\\ 0 & \text{else,}\end{array}\right.
\end{equation}
such that features with an absolute correlation below $\tau$ are not penalized, while higher absolute correlations are assigned penalties that represent the level of correlation. As a result, the selected feature set contains features with lower mutual correlations.\footnote{We suggest to use Spearman's rho as correlation coefficient, since it is robust (in contrast to Pearson's correlation coefficient) and faster to compute than Kendall's tau.}

\paragraph{Feature block priors}
User knowledge may as well be available for \textit{feature blocks} rather than for single features. Feature blocks are contextual groups of features, such as those extracted from the same source in a multi-source dataset. It can be desirable to select features from a few distinct blocks so that the model does not depend on all sources at once. While prior weights can be trivially assigned on block level, we transfer the concept of side constraints to feature blocks.

Feature blocks are specified via a block matrix $\bm{B} \in \{0,1\}^{W\times N}$, where $1$ indicates that the feature $n\in [N]$ is part of block $w\in[W]$ and $0$, else. Even though a full partition of the feature set is common, feature blocks are neither required to be mutually exclusive, nor exhaustive. Along with the block matrix $\bm{B}$, an inequality system between blocks consists of a matrix $\bm{A}^{\text{block}}\in\mathbb{R}^{K\times W}$ and a vector $\bm{b}^{\text{block}}\in\mathbb{R}^{K}$. To evaluate whether a block is selected by a feature set $\bm{\delta}$, we define the block selection vector $\bm{\delta}^{\text{block}}\in \{0,1\}^{W}$, given by
\begin{equation}
\bm{\delta}^{\text{block}} = \left(\bm{B}\bm{\delta}\geq \mathbbm{1}_W\right), 
\end{equation}
where $\geq$ refers to an element-wise comparison of vectors, delivering 1 for a component, if the condition is fulfilled, and 0, otherwise. In other words, a feature block is selected, if at least one feature of the corresponding block is selected. Although block constraints introduce non-linearity into the system of side constraints, they can be used in the same way as linear constraints between features and integrated into the joint admissibility function $\kappa$.

\subsection{Optimization}

Exploiting the conjugate prior property, the posterior density of $\bm{\theta}$ can be expressed as a Dirichlet, generalized Dirichlet or hyperdirichlet distribution, respectively. Since the expected value $\mathbb{E}_{\bm{\theta}}[\bm{\theta}]$ can be computed either in a closed-form expression (Dirichlet or generalized Dirichlet)~\cite{wong98}, or simulated via a sampling procedure (hyperdirichlet)~\cite{hankin}, it remains to solve the discrete optimization problem in Eq. \ref{eq:loss} as a final step.

\begin{algorithm}
\caption{Probabilistic sampling algorithm to initialize GA.}
  \label{alg:sampling2}
\begin{algorithmic}
\Require $\bm{\alpha}^{\circ}$, $\bm{A}$, $\bm{b}$, $\bm{\rho}$, sample size $Q$
\State $G \leftarrow \{\}$\;
\For{$q\in [Q]$}
    \State $\bm{\delta} \leftarrow (0,0,\dots,0)$\;
    \State generate a permutation $\pi$ on $[N]$ by sampling $N$ times without replacement with probabilities proportional to $\bm{\alpha}^\circ$\;
    \For{$i = \pi(1),\dots,\pi(N)$}
    	\State define $\bm{\delta}^{\dagger}$ as 
    		$\delta^{\dagger}_n \leftarrow \left\{
    			\begin{array}{l l}     
    				\delta_n & n\neq i\\
        			1 & n = i
        		\end{array}\right.$ 
        	for each $n\in [N]$\;
        \State sample $u \sim \text{Unif}_{[0,1]}$\;
        \If{$u \leq r_{\bm{\delta}^{\dagger},\bm{\delta}}$}
            \State update $\bm{\delta} \leftarrow \bm{\delta}^{\dagger}$\;
        \EndIf
    \EndFor
	\State $G \leftarrow G \cup \{\bm{\delta}\}$\;
\EndFor \\
\Return $G$
\end{algorithmic}
\end{algorithm}

Since an analytical minimization is not feasible, we determine a numerical optimum $\bm{\delta}^{\star}$ by using discrete optimization: we deploy the Genetic Algorithm (GA) described by \cite{givens:compstat}. 
To guarantee a fast convergence towards an acceptable solution, it is beneficial to provide initial samples, which are good candidates for the final solution. For this purpose we propose a probabilistic sampling algorithm, Alg. \ref{alg:sampling2}: 
In essence, the algorithm creates a random permutation of all features, $\pi:[N]\rightarrow[N]$, by weighted and ordered sampling without replacement. The weights represent the posterior parameter vector $\bm{\alpha}^{\circ}$. Then, the algorithm iteratively accepts or rejects feature $\pi(n)$ with a success probability
\begin{equation}
    r_{\bm{\delta}^{\dagger},\bm{\delta}} = \left\{\begin{array}{l l} \frac{\kappa(\bm{\delta}^{\dagger})}{\kappa(\bm{\delta})} & \text{if}~\kappa(\bm{\delta}) > 0 \\
    0 & \text{else,}
    \end{array}\right.
\end{equation}
denoting the admissibility ratios of feature sets with and without feature $\pi(n)$. The generated sample accounts for high feature weights by low ranks, resulting in a higher probability to be accepted in the acceptance/rejection step.

The Genetic Algorithm (GA) for discrete optimization is initialized using Algorithm \ref{alg:sampling2}.
Starting with an initial set of feature membership vectors $\left\{\bm{\delta}^{0}\in\{0,1\}^N\right\}$, GA creates new vectors $\bm{\delta}^{t}\in\{0,1\}^N$ as pairwise combinations of two preceding vectors $\bm{\delta}^{t-1}$ and $\tilde{\bm{\delta}}^{t-1}$ in each iteration $t\in[T]$. A combination refers to sampling component $\bm{\delta}^{t}_n$ from either $\bm{\delta}^{t-1}_n$ or $\tilde{\bm{\delta}}^{t-1}_n$ in a uniform way and adding minor random mutations to single components. The posterior density serves as fitness when deciding which vectors $\bm{\delta}^{t-1}$ and $\tilde{\bm{\delta}}^{t-1}$ from iteration $t-1$ should be combined to $\bm{\delta}^{t}$ --- the fitter, the more likely to be part of a combination. 

\section{Experiments \& Results}
\label{sec:experiments}

Our experiments evaluate the performance, flexibility, and applicability of UBayFS in two parts: first, a study conducted on synthetic datasets demonstrates the properties of the various model parameters, including
\begin{enumerate}
    \item[a.] the number of elementary models $M$ (1a),
    \item[b.] the prior weights $\bm{\alpha}$ in a block-wise setup (1b),
    \item[c.] the constraint types and their shapes $\rho$ in a block-wise setup (1c), as well as
    \item[d.] the type of prior distribution to account for feature dependencies (1d).
\end{enumerate}
The second part of the experiment is conducted on real-world classification datasets from the life science domain. We demonstrate the advantageous quality of the UBayFS framework in comparison with state-of-the-art ensemble feature selectors. The experiment also includes a block feature selection setup for datasets with block structure. 

\paragraph{Default parameters} In all of these synthetic experiments, six elementary feature selectors with different complexities are used:
\begin{itemize}
    \item minimum Redundancy Maximum Relevance (mRMR) \cite{ding:mrmr},
    \item Fisher score \cite{bishop:fisherscore},
    \item decision tree for classification \cite{decision_trees},
    \item recursive feature elimination (RFE) \cite{Guyon2002:rfe},
    \item Hilbert-Schmidt Independence Criterion Lasso (HSIC) \cite{yamada14},
    \item Lasso \cite{tibshirani:lasso}.
\end{itemize}
Nevertheless, the main focus of the present work is to demonstrate the merits of the generic concept of UBayFS rather than to provide an in-depth analysis of the elementary feature selectors.

Our implementation of UBayFS\footnote{An implementation in R is publicly available at \url{https://github.com/annajenul/UBayFS}. Experimental setups are provided at \url{https://github.com/annajenul/UBayFS_experiments}. For details on the datasets, see Appendix \ref{secA2}.} in R (\cite{R}) uses the Genetic Algorithm package authored by \cite{R:GA} with $T=100$ and $Q = 100$---in most cases, the optimum is reached after around ten iterations. By default, each UBayFS setup comprises an uninformative prior with $\alpha_n=0.01$ for all $n\in[N]$, and a max-size constraint instructing to select $b_{\text{MS}}$ features, which is determined individually for each dataset. Each setup is executed in $I = 10$ independent runs $i \in [I]$, representing distinct random splits of the dataset $\mathcal{D}$ into train data $T_{\text{train}}^{(i)}$ and test data $T_{\text{test}}^{(i)} = \mathcal{D}\setminus T_{\text{train}}^{(i)}$ (stratified 75\%/25\% split). The feature selector is applied on $T_{\text{train}}^{(i)}$ and predictive performances are evaluated on $T_{\text{test}}^{(i)}$.

\paragraph{Evaluation metrics} For the synthetic datasets, performance is measured by the F1 score of correctly / incorrectly selected features since the ground truth about the relevance of features is known from the simulation procedure. For real-world data, F1 scores on the predictive results are used to judge the feature selection quality indirectly. Furthermore, all experiments use the \textit{stability} measure by \cite{nogueira:stability} to assess the agreement between results from $I$ independent feature selection runs. Stability ranges asymptotically in $[0,1]$, where $1$ indicates that the same features are selected in every run (perfectly stable). \textit{Runtime}\footnote{CentOS Linux 7.9.2009, Intel Xeon(R) CPU E5-2650 @ 2.60GHz, 3 GB RAM, R v3.6.0.} refers to the time the model requires to perform feature selection, including elementary model training and optimization, but excluding any predictive model trained on top of the feature selection results. Since prior parameters have a minor influence on the runtime, times will not be provided for experiments investigating these aspects.

\subsection{Experiment 1: Simulation study}

In the simulated case, we make use of three different datasets:
\begin{enumerate}
    \item[i.] an additive model (experiment 1a) similar to \textit{Data1} in \cite{yamada14}, composed of a $(x_1,\dots,x_{1000})\sim 1000\times 1000$ data matrix simulated from a Gaussian distribution $N(\bm{0}_{1000},\bm{I}_{1000})$, and a target variable for classification, given by $$y=g(-2\sin(2x_1)+x_2^2+x_3+\exp(-x_4)+\varepsilon),$$ where $x_1,\dots,x_4$ denote the features $1$ to $4$ and $\varepsilon \sim N(0,1)$. The function $g$ transforms $z$ into a class variable by
    $$g(z)=\left\{\begin{array}{ll}
        1 & \text{if}~z\geq 0,\\
        0 & \text{otherwise;}
    \end{array}\right.$$
    \item[ii.] a non-additive model (experiment 1a) similar to \textit{Data2} in \cite{yamada14}, equivalent to the setup of i., except for a target variable $$y=g(x_1\cdot \exp(2x_2)+x_3^2+\varepsilon);$$
    \item[iii.] a simulated dataset  (experiment 1b, 1c) with group structure among the features, produced via \textit{make\_classification} \cite{scikit-learn}, delivering a $512\times 256$ dataset with $8$ features blocks à $32$ features---4 of these blocks contain relevant features (4 important features per block), 2 blocks contain redundant features representing arbitrary linear combinations of the relevant features (3 redundant features per block);
    \item[iv.] another dataset simulated via \textit{make\_classification}, comprising 32 features in total (16 important, 16 redundant) without block structure. This smaller dataset ($64\times 32$) has a complicated correlation structure due to the high number of redundant features and is used to evaluate UBayFS variants that take feature dependence into account (experiment 1d).
\end{enumerate}
The maximum number of selected features $b_{\text{MS}}$ is set to the ground truth number of relevant features, i.e. $b_\text{MS}=4$ (dataset i.), $b_\text{MS}=3$ (dataset ii.), and $b_\text{MS}=16$ (dataset iii.), respectively. The default constraint shape parameters for MS is set to $\rho_{\text{MS}} = 1$. Unless otherwise stated, the prior weights are set to a constant, uninformative value of $\alpha=0.01$ for all features.

In addition to the constraint shape $\rho$ associated with a single constraint, $\lambda$ balances the overall impact of side constraints with the Dirichlet-multinomial model. However, a small parameter $\lambda<1$ is not recommended since a lack of influential constraints (including the MS constraint) results in selecting all features due to a monotonic target function. On the other hand, a high $\lambda$ has a similar effect as setting all shape parameters uniformly to $\rho=\infty$; thus, all constraints are required to be fulfilled. In this study, $\lambda$ does not significantly impact the resulting model metrics and, therefore, is set to $\lambda=1$ and not further evaluated in this study.

\begin{figure}[ht]
    \centering
    \subfigure[additive classification dataset]{
    \includegraphics[width=0.6\textwidth]{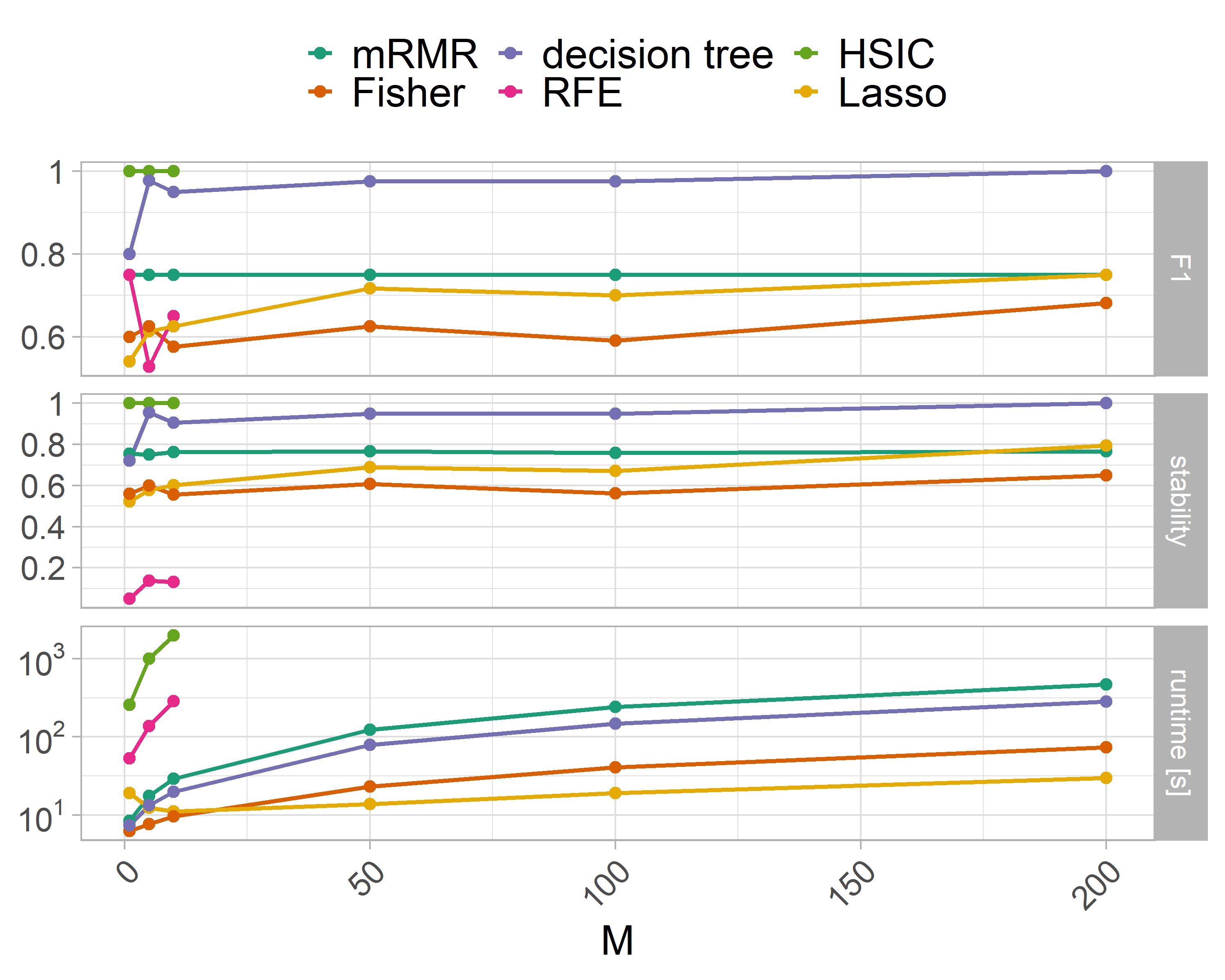}
    }
    \subfigure[non-additive classification dataset]{
    \includegraphics[width=0.6\textwidth]{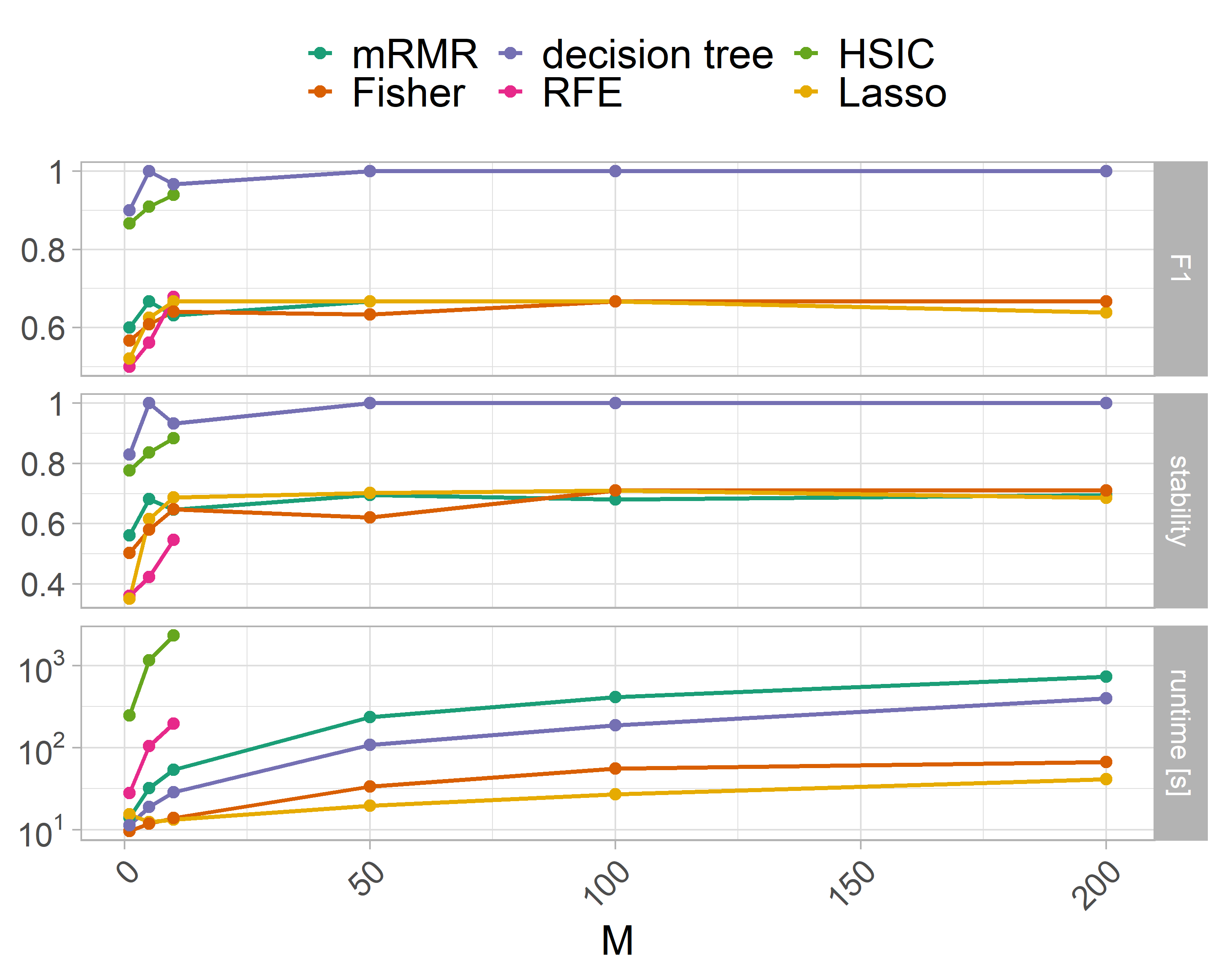}
    }
    \caption{Different numbers of elementary models $M$.}
    \label{fig:ex1a}
\end{figure}

\paragraph{Experiment 1a---likelihood parameters}
Fig. \ref{fig:ex1a} demonstrates the effect of an increasing number of elementary models $M$ to build the feature selector. Along with the choice of the elementary feature selector, $M$ represents the parameter to steer the likelihood. Due to their excessive runtimes, HSIC and RFE are computed only for $M\leq 10$, while all other elementary feature selectors are evaluated for up to $M=200$.

As expected, a higher $M$ contributes largely to the runtime of the model, which increases linearly. In contrast, both F1 scores and stability values begin to saturate at around $M=50$ to $M=100$ models. Even though large ensembles are intractable with HSIC and RFE, small ensembles with $M=5$ allow HSIC to retrieve almost all features, whereas simpler elementary feature selectors struggle to achieve high performances and stabilities even at higher levels of $M$. We conclude that large $M$ does not necessarily improve the results, but significantly impacts the runtime, thus $M\approx 100$ appears to be a reasonable choice in the subsequent settings, except for HSIC and RFE, where $M=5$ will be set as a default.

\paragraph{Experiment 1b---block-wise prior weights}
To investigate the effect of prior weights, we alter the prior weights for the four blocks containing relevant features (according to the simulation of dataset iii.). A constant prior weight $\alpha_R$ is assigned to all features from relevant blocks, i.e., block containing relevant features. In contrast, features from all other blocks are assigned a constant prior weight $\alpha_{-R}$---thereby, we simulate that the expert has approximate, yet not exact beliefs about features relevance. By assigning higher prior weights $\alpha_R>\alpha_{-R}$, the experiment represents an agreement between the expert belief and the ground truth, while a lower $\alpha_{R}<\alpha_{-R}$ represents "wrong" prior information. In this experiment, we alternatively increase either $\alpha_{R}$ or $\alpha_{-R}$ while setting the other to the default value $0.01$.

\begin{figure}[ht]
    \centering
    \includegraphics[width=0.6\textwidth]{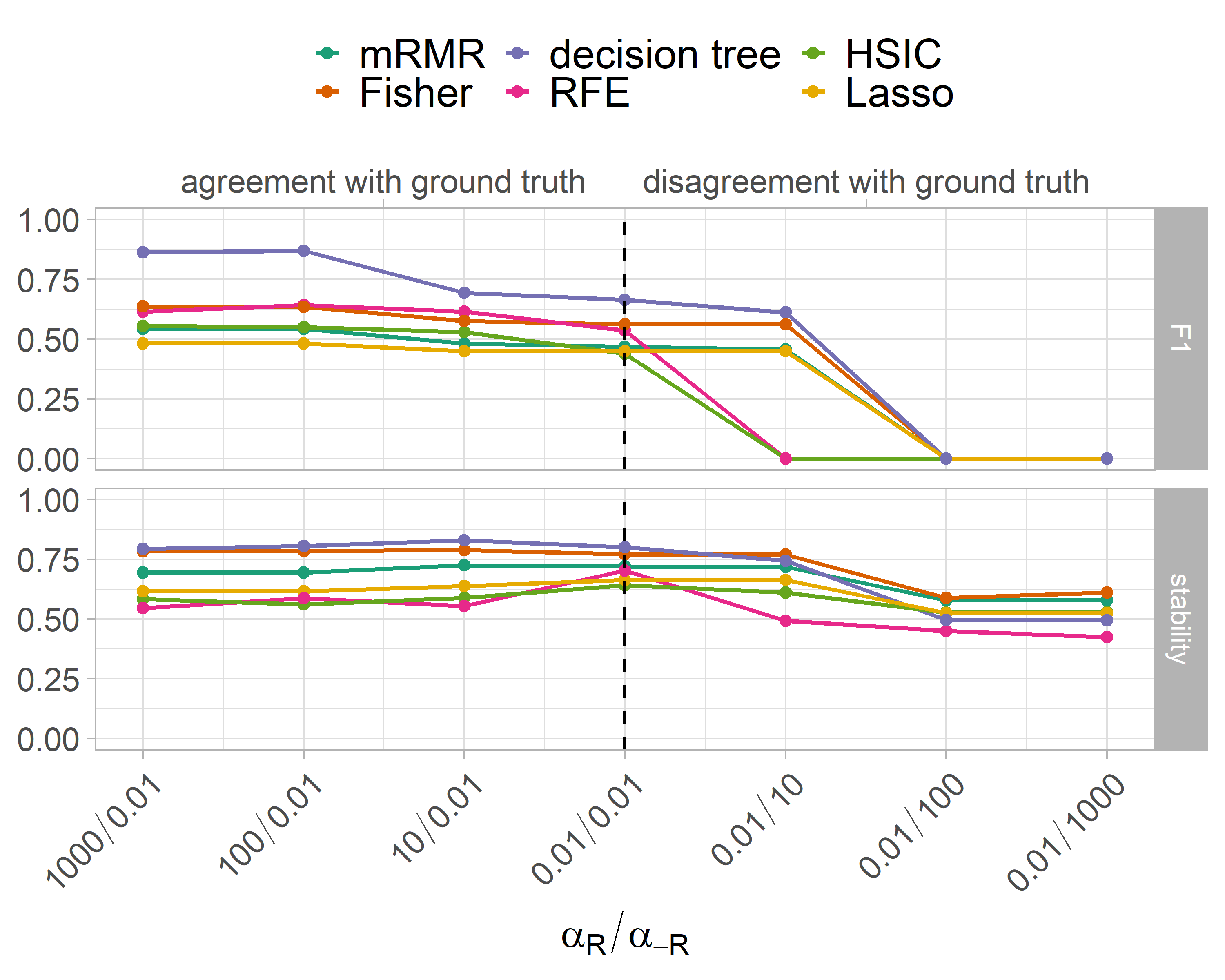}
    \caption{Different prior weights assigned to relevant blocks, $\alpha_R$, and to non-relevant blocks, $\alpha_{-R}$.}
   \label{fig:ex1b}
\end{figure}

Fig. \ref{fig:ex1b} illustrates that, as expected, feature selection performance in terms of F1 scores (evaluated with respect to the ground truth features) increases for higher $\alpha_R$ and decreases for higher $\alpha_{-R}$. Thus, across all elementary feature selectors, an improvement of the uninformative case $\alpha_{R}=\alpha_{-R}=0.01$ can be achieved by an informative prior, if the prior represents a reasonable overlap with reality---this holds even though the relevant block also contain uninformative features, which are incremented by $\alpha_R$ as well. On the other hand, erroneous prior knowledge can impact the feature selection results negatively. In contrast to the feature-wise F1 scores, stability remains mostly unaffected from strong prior knowledge on relevant or irrelevant blocks---incorrect prior knowledge merely tends to decrease stability to a minor degree.

\paragraph{Experiment 1c---block constraints}
We investigate the following opposite constraint types:
\begin{itemize}
    \item \textit{block-max-size} (BMS): (soft) upper limit $b_{\text{BMS}}$ to the number of selected blocks, and
    \item \textit{max-per-block} (MPB): at most $b_{\text{MPB}}$ features can be selected from the same block.
\end{itemize}
BMS is designed to enforce a clustering behavior, where all features (at most $b_{\text{MS}}=16$, according to the MS constraint) are selected from a maximum number of $b_{\text{BMS}} = 4$ blocks. On the other hand, MPB aims to disperse the selection, indicating that a maximum number of $b_{\text{MPB}}=2$ feature per blocks is favorable. The strength of these constraints is steered via the according shape parameters $\rho_{\text{BMS}}$ and $\rho_{\text{MPB}}$, respectively. Per default, we indicate $\rho = 0$ in cases where a constraint is omitted. From a default case of $\rho_{\text{BMS}}=\rho_{MPB}=0$ (no block constraints), we investigate the behavior of UBayFS in both directions, i.e. for an increasing level of $\rho_{\text{BMS}}$ or $\rho_{MPB}$.

Fig. \ref{fig:ex1b} illustrates how the opposite prior constraints BMS and MPB affect the model at different levels of relaxation parameters. Both constraint types have a slightly negative impact on the outcome in terms of F1 and stability. This is caused by the fact that the "best" feature set has to be determined under a side constraint, which is not compatible with the ground truth---the ground truth defines 16 features out of four distinct blocks to be relevant, which cannot be covered by any of the constraints. Therefore, we can observe that UBayFS can handle such scenarios and still deliver appropriate and near-optimal solutions.

\begin{figure}[ht]
    \centering
    \includegraphics[width=0.6\textwidth]{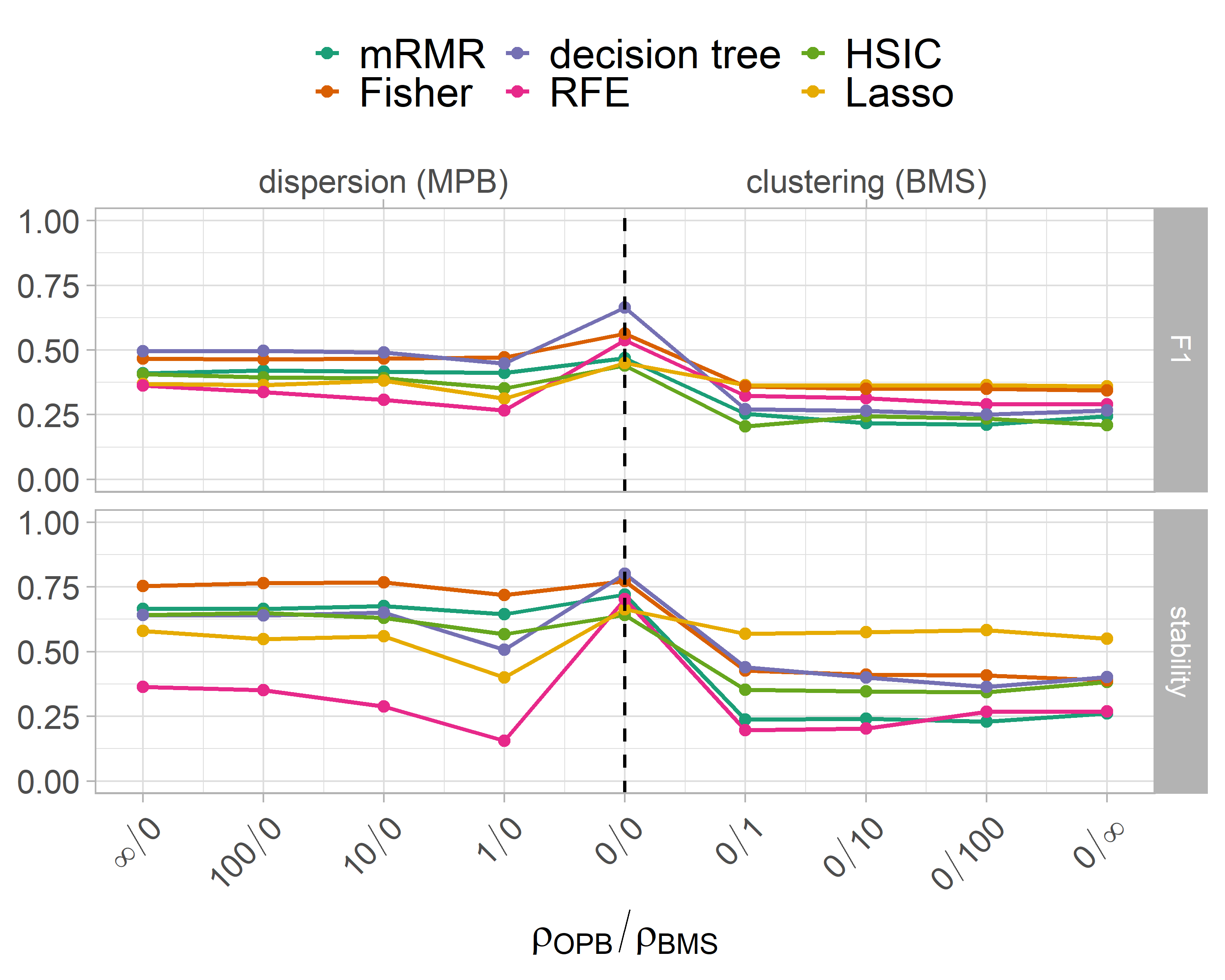}
    \caption{Different prior constraint assigned to blocks: MPB (maximum one feature per block) and BMS (block max-size) constraint types at distinct levels of $\rho$. The special case $\rho=0$ indicates that the according constraint is omitted.}
   \label{fig:ex1c}
\end{figure}

\paragraph{Experiment 1d---feature dependence models}
In Section \ref{sec:method}, multiple variants were discussed to account for datasets with correlation structure. On the one hand, the UBayFS framework permits to account for between-data correlations via a generalization of the prior distribution; on the other hand, we may enforce that the highly correlated features should not be selected jointly via a decorrelation constraint. Both variants are different insofar as generalized priors aim to deliver a more appropriate estimation of the expected feature importances by correcting for dependencies in the observed feature sets, while decorrelation constraints directly affect the optimization procedure for $\bm{\delta}$.

In this experiment, we investigate both possibilities to account for dependencies between features, along with combinations of both: we set a decorrelation constraint between all features with a mutual Spearman correlation $\tau>0.4$ as described in Section \ref{sec:side_constraints}. Generalizations of the Dirichlet prior setup are denoted as follows:
\begin{itemize}
    \item Dirichlet prior distribution,
    \item generalized Dirichlet distribution \cite{wong98},
    \item hyperdirichlet distribution \cite{hankin}.
\end{itemize}
Our experiment involves all combinations of prior setups with and without decorrelation constraint, executed on dataset vi. To measure the effect of decorrelation, we further evaluate the redundancy rate (RED) as suggested in \cite{zhao10}: the redundancy rate of a feature set is defined as the average absolute Pearson correlation between all pairs of distinct features in the selected feature set. A small RED is preferred in many practical setups.

The results show that neither feature-wise F1 scores, nor stabilities change significantly between the prior models. Thus, the default Dirichlet model seems sufficient to obtain reasonable results. However, introducing decorrelation constraints has a slightly negative impact on stability, while yielding a small improvement in F1 scores and RED. Nonetheless, the most significant change between the variants can be observed with respect to runtime, which reflects the high computational burden associated with the hyperdirichlet prior model---even on a small dataset, the runtimes show a significant increase on a logarithmic scale. Thus, higher-dimensional datasets cannot be tackled with the hyperdirichlet setup.

\begin{figure}[ht]
    \centering
    \includegraphics[width=0.6\textwidth]{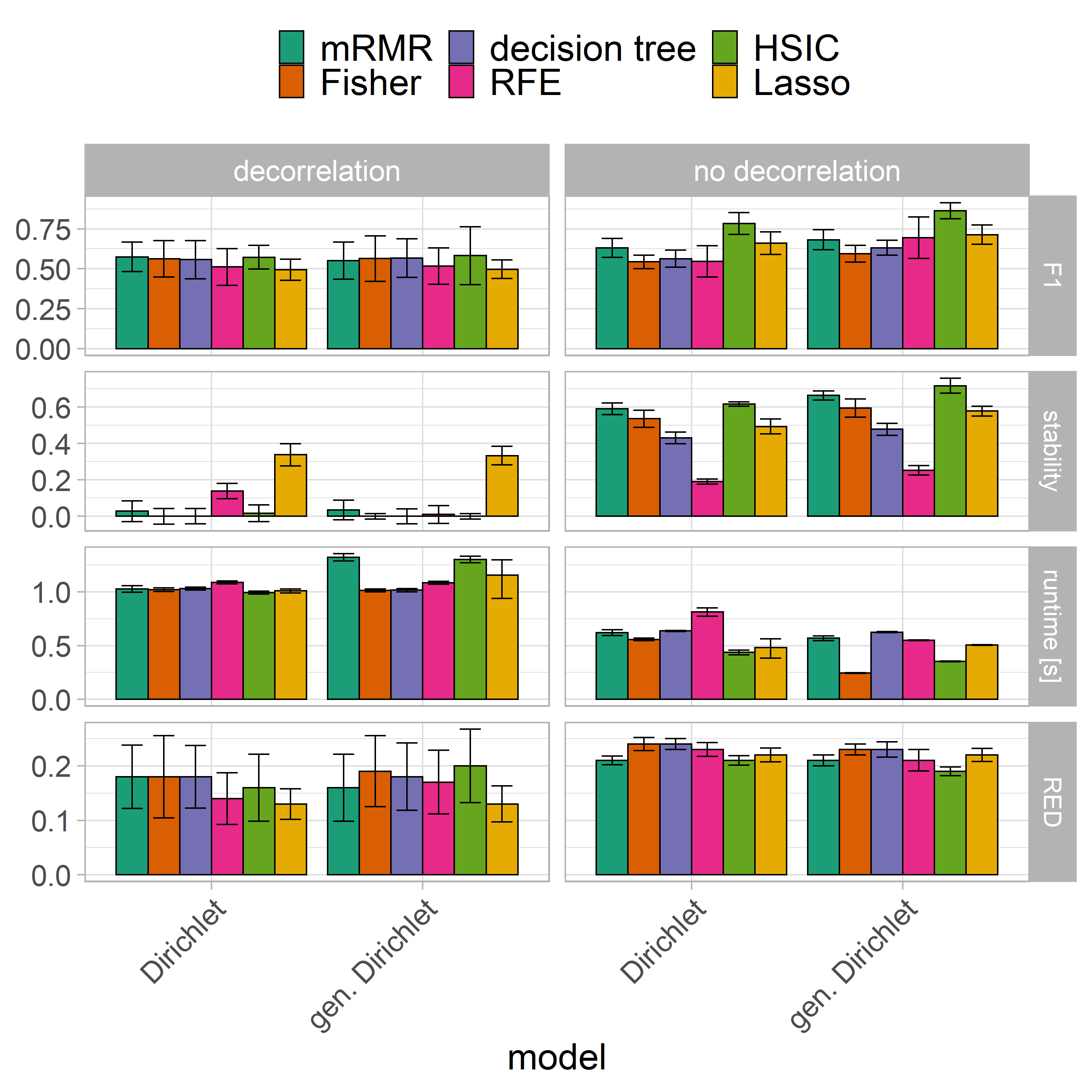}
    \caption{Different setups to account for dependence structures between features.}
   \label{fig:ex1d}
\end{figure}

\subsection{Experiment 2: Real-world life sciences datasets}

Real-world experiments are conducted on seven open-source datasets presenting binary classification problems from the life science domain, see Tab. \ref{tab:datasets}. For simplicity and due to extensive runtimes, we restrict the choice of the elementary feature selector for UBayFS to mRMR, Fisher, and decision tree with an uninformative prior, an MS constraint, and $M=100$. The number of selected features is specified according to the size of the dataset ($b_\text{MS}=5$ / $10$ / $20$ for datasets with fewer than 100 / between 100 and 1000 / more than 1000 features, respectively). 

\begin{table}[ht]
    \centering    
    \caption{Real-world binary classification datasets from the life science domain used for experimental evaluation. For p53, a stratified subset out of $>16000$ rows was used from the original dataset for this experiment.}
    \label{tab:datasets}
    \begin{tabular}{lrrrrr}
        \toprule
        dataset source & \# features & \# blocks & \# rows & $b_{MS}$ & $b_{BMS}$  \\
        \midrule
        Breast Cancer Wisconsin (BCW) \cite{wolberg:wisconsin} & 30 & 3 & 569 & 5 & 1 \\
        Heart Disease (HD) \cite{detrano:heart} & 46 & 1 & 101 & 5 & -\\
        Mice Protein Expression (MPE) \cite{higuera:mouse} & 77 & 1 & 552 & 5 & - \\
        Colon Gene Expression (COL) \cite{yang:col} & 100 & 20 & 62 & 5 & 2 \\
        LSVT Voice Rehabilitation \cite{tsanas:LSVT} & 310 & 14 & 126 & 10 & 2 \\
        p53 \cite{danziger:p53} & 5409 & 2 & 351 & 20 & 1 \\
        Prostate (PRO) \cite{pro02} & 6033 & 1 & 102 & 20 & - \\
        Leukaemia (LEU) \cite{leu99} & 7129 & 1 & 72 & 20 & - \\
        \bottomrule
    \end{tabular}

\end{table}

We evaluate two different scenarios in this experiment: scenario 1 is a standard feature selection scenario, where a number of $b_{\text{MS}}$ features (specified in Tab. \ref{tab:datasets}) should be selected. Scenario 2 applies only to datasets with block structure, i.e., more than one block, and evaluates block feature selection: a number of up to $b_{\text{MS}}$ features should be selected from at most $b_{\text{BMS}}$ distinct blocks.\footnote{Details on the blocks are provided in Appendix \ref{secA2}.} Random forests (RF) \cite{br:randomforest}, and RENT \cite{jenul:rent} (representing ensemble feature selectors that extend the concepts of decision trees and elastic net regularized models, respectively) are used as state-of-the-art benchmarks for standard feature selection, while Sparse Group Lasso (GL) \cite{ida:fSparseGlasso} is used as the benchmark for block feature selection. To conform with UBayFS, RENT and RF are adjusted to $M=100$ elementary models, and all models are tuned to select approximately the same number of features, $b_{\text{MS}}$. Since RENT and GL cannot be instructed to select $b_{\text{MS}}$ features directly, regularization parameters are determined via bisection, such that the number of selected features is approximately equal to $b_{\text{MS}}$.

The selected features cannot be evaluated directly in real-world datasets due to unknown ground truth on the feature relevance. Therefore, we train predictive models on $T_{\text{train}}^{(i)}$ after feature selection and evaluate the selected features indirectly via the predictive performance on the test instances. To reduce the influence of the predictive model type, we train a two distinct classifiers on $T_{\text{train}}^{(i)}$ after feature selection, and report F1 scores for predictions on $T_{\text{test}}^{(i)}$ for both. The choice of baseline classifiers to obtain the prediction comprises:
\begin{itemize}
    \item generalized linear model: logistic regression (GLM),
    \item support vector machine (SVM).
\end{itemize}

\begin{table}[t]
    \caption{UBayFS with three distinct elementary feature selectors (M: mRMR, F: Fisher, T: decision tree) is compared to ensemble feature selectors RF and RENT in a standard feature selection scenario. Further, UBayFS with additional (BMS) constraint is compared to Sparse Group Lasso (GL) for block-feature selection on datasets with block structure. Average F1 scores are given for different predictive models (GLM, SVM). The best scores in each row are marked in bold for each scenario.}
    \label{tab:exp2_results}
    \centering
    \subtable[Average F1 score per run (predictor: GLM).]{
    \begin{tabular}{lrrrrr|rrrr}
        \toprule
        \multirow{3}{*}{dataset} & \multicolumn{5}{c|}{standard feature selection} & \multicolumn{4}{c}{block feature selection}\\ \cmidrule{2-10}
        & \multirow{2}{*}{RF} & \multirow{2}{*}{RENT} & \multicolumn{3}{c|}{UBayFS}& \multirow{2}{*}{GL} & \multicolumn{3}{c}{UBayFS}\\
        & & & M & F & T & & M & F & T\\
        \midrule
        BCW    & 0.95 & $\bm{0.97}$ & 0.96 & $\bm{0.97}$   & 0.95  & $\bm{0.96}$ & $\bm{0.96}$ & $\bm{0.96}$   & $\bm{0.96}$  \\
        HD     & 0.92 & 0.88 & 0.91 & 0.90   & $\bm{0.93}$  & - & - & - & - \\
        MPE    & 0.86 & $\bm{0.95}$ & 0.87 & 0.83   & 0.83  & - & - & - & - \\
        COL    & 0.85 & 0.83 & 0.83 & 0.78   & $\bm{0.88}$  & 0.82 & 0.74 & 0.77   & $\bm{0.89}$  \\
        LSVT   & 0.70 & 0.75 & 0.80 & $\bm{0.84}$   & 0.68  & 0.77 & 0.67 & $\bm{0.79}$   & 0.59  \\
        p53    & 0.71 & 0.66 & $\bm{0.80}$ & 0.78 & $\bm{0.80}$ & 0.63 & 0.76 & $\bm{0.79}$ & $\bm{0.79}$ \\
        PRO    & 0.88 & $\bm{0.89}$ & 0.78 & 0.85   & 0.84  & - & - & - & - \\
        LEU   & 0.88 & 0.93 & 0.88 & 0.91   & $\bm{0.95}$  & - & - & - & - \\
        \bottomrule
        \end{tabular}
    }
    \subtable[Average F1 score per run (predictor: SVM).]{
    \begin{tabular}{lrrrrr|rrrr}
        \toprule
        \multirow{3}{*}{dataset} & \multicolumn{5}{c|}{standard feature selection} & \multicolumn{4}{c}{block feature selection}\\ \cmidrule{2-10}
        & \multirow{2}{*}{RF} & \multirow{2}{*}{RENT} & \multicolumn{3}{c|}{UBayFS}& \multirow{2}{*}{GL} & \multicolumn{3}{c}{UBayFS}\\
        & & & M & F & T & & M & F & T\\
        \midrule
        BCW    & 0.95 & $\bm{0.97}$ & 0.96 & 0.96   & 0.94  & $\bm{0.97}$ & 0.96 & 0.96   & 0.95  \\
        HD     & 0.92 & 0.88 & 0.91 & 0.91   & $\bm{0.95}$  & - & - & - & - \\
        MPE    & 0.87 & $\bm{0.95}$ & 0.89 & 0.84   & 0.84  & - & - & - & - \\
        COL    & 0.86 & 0.85 & 0.87 & 0.83   & $\bm{0.88}$  & 0.81 & 0.82 & 0.79   & $\bm{0.89}$  \\
        LSVT   & 0.75 & 0.75 & 0.80 & $\bm{0.84}$   & 0.71  & $\bm{0.80}$ & 0.79 & 0.79   & 0.57  \\
        p53    & 0.81 & $\bm{0.82}$  & 0.81 & 0.80 & $\bm{0.82}$  & $\bm{0.84}$ & 0.77 & 0.82   & 0.80  \\
        PRO    & $\bm{0.91}$ & 0.90 & 0.87 & 0.88   & 0.85  & - & - & - & - \\
        LEU   & $\bm{0.96}$ & 0.94 & 0.88 & 0.95   & $\bm{0.96}$  & - & - & - & - \\
        \bottomrule
        \end{tabular}
    }
\end{table}
\begin{table}[t]
    \caption{Mean stabilities of UBayFS with three distinct elementary feature selectors (M: mRMR, F: Fisher, T: decision tree), compared to ensemble feature selectors RF and RENT in standard feature selection, as well as to GL in block feature selection scenarios. The best scores in each row are marked in bold for each scenario.}
    \label{tab:exp2_results_stab}
    \centering
    \begin{tabular}{lrrrrr|rrrr}
        \toprule
        \multirow{3}{*}{dataset} & \multicolumn{5}{c|}{standard feature selection} & \multicolumn{4}{c}{block feature selection}\\ \cmidrule{2-10}
        & \multirow{2}{*}{RF} & \multirow{2}{*}{RENT} & \multicolumn{3}{c|}{UBayFS}& \multirow{2}{*}{GL} & \multicolumn{3}{c}{UBayFS}\\
        & & & M & F & T & & M & F & T\\
        \midrule
        BCW    & 0.73 & 0.87 & 0.87 & $\bm{1.00}$   & 0.61  & $\bm{0.90}$ & 0.80 & 0.80   & 0.80  \\
        HD     & 0.45 & 0.87 & $\bm{0.88}$ & 0.65   & 0.59  & - & - & - & - \\
        MPE    & 0.72 & $\bm{0.87}$ & 0.92 & 0.85   & 0.77  & - & - & - & - \\
        COL    & 0.39 & 0.67 & 0.80 & 0.72   & $\bm{0.81}$  & 0.56 & $\bm{0.84}$ & 0.72   & 0.82  \\
        LSVT   & 0.31 & 0.59 & 0.72 & $\bm{0.79}$   & 0.55  & 0.73 & 0.66 & $\bm{0.88}$   & 0.31  \\
        p53    & 0.11 & $\bm{0.56}$ & 0.34 & 0.34 & 0.36 & $\bm{0.68}$ & 0.19 & 0.25 & 0.31 \\
        PRO    & 0.17 & 0.53 & 0.56 & $\bm{0.61}$   & 0.42  & - & - & - & - \\
        LEU   & 0.07 & 0.64 & 0.46 & $\bm{0.76}$   & 0.53  & - & - & - & - \\
        \bottomrule
        \end{tabular}
\end{table}

\begin{figure}[t]
    \centering
    \includegraphics[width=0.6\textwidth]{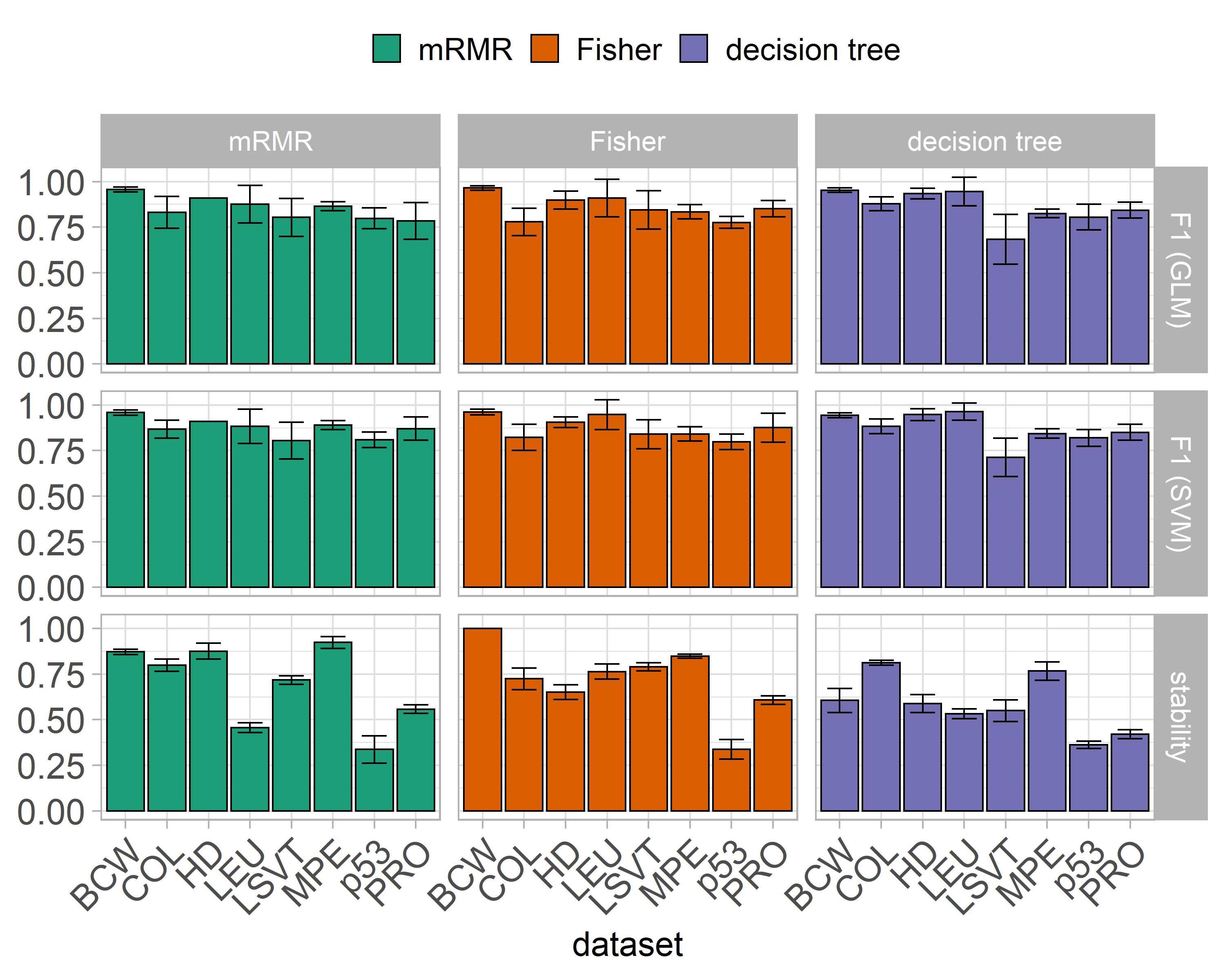}
    \caption{Performance results of UBayFS feature selection on real-world datasets (MS constraint). F1 scores are determined after training and predicting a classifier (GLM or SVM) after feature selection. Results show mean values over $I=10$ runs along with standard deviations.}
    \label{fig:exp2}
\end{figure}

\begin{figure}[t]
    \centering
    \includegraphics[width=0.6\textwidth]{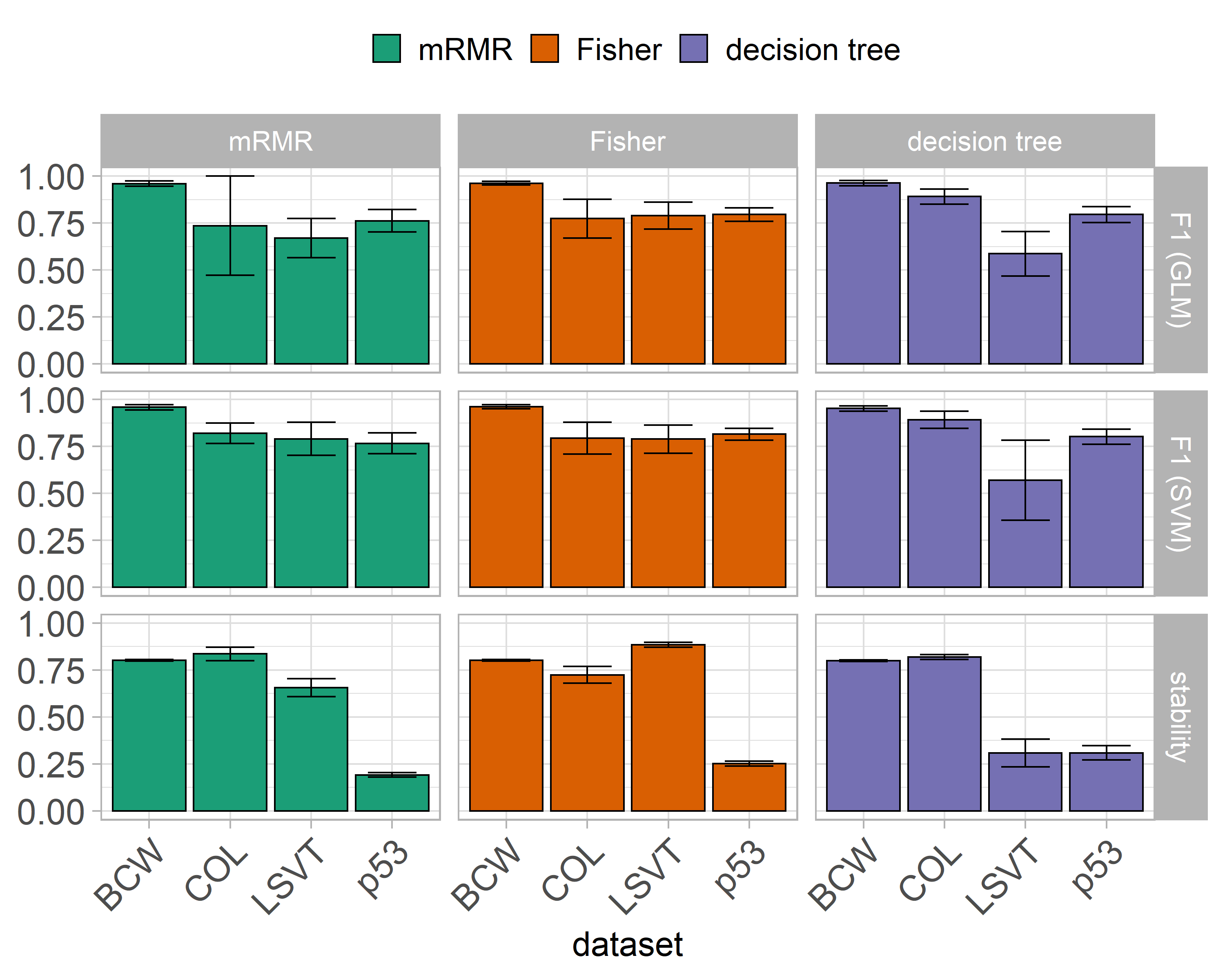}
    \caption{Performance results of UBayFS block feature selection on real-world datasets (MS and BMS constraints). F1 scores are determined after training and predicting a classifier (GLM or SVM) after feature selection. Results show mean values over $I=10$ runs along with standard deviations.}
    \label{fig:exp2_block}
\end{figure}

\paragraph{Results} 
Tab. \ref{tab:exp2_results} and \ref{tab:exp2_results_stab} present the results of the experiments on real-world data. Thereby, UBayFS can keep up with other approaches and achieves good predictive F1 scores throughout the different datasets, even though only a limited amount of expert knowledge is introduced to ensure a fair comparison. In the block feature selection setups, UBayFS benefits from block constraints and shows more flexibility than Sparse Group Lasso. Altogether, F1 scores are generally in a high range across all methods, suggesting that UBayFS can keep up or even outperform its competitors in a diverse range of scenarios (low-dimensional and high-dimensional data, as well as unconstrained and constrained setups). Fig. \ref{fig:exp2} and Fig. \ref{fig:exp2_block} provide additional insights into the performances of the UBayFS variants in the standard feature selection and block feature selection scenario, respectively.

Overall, the results reflect that a particular strength of UBayFS lies in delivering a good trade-off between stabilities and predictive performance, compared to competitors like RF, which deliver high F1 scores, but very low stabilities. Differences between the F1 scores obtained by the different elementary feature selectors underline that UBayFS inherits benefits and drawbacks from its underlying elementary model type---in particular, the decision tree and HSIC achieved top results. Nevertheless, the building of ensembles allows to compensate in parts for mediocre stabilities.

Runtimes of all methods and datasets are provided in Tab. \ref{tab:runtimes}. Given a fixed set of model parameters, it becomes obvious that the major factor influencing the runtime of UBayFS is the number of features (columns) rather than the number of samples (rows). UBayFS runtimes refer to the MS setup---however, experiments showed only minor differences to the runtimes in the block feature selection setup. While RF and GL are more tractable in high-dimensional datasets, RENT seems to suffer from data dimensionality to a more considerable extent.

\begin{table}[t]
    \centering
    \caption{Average runtime per run [s].
     \label{tab:runtimes}}
     \begin{tabular}{lrrrrrr}
        \toprule
        \multirow{2}{*}{dataset} & \multirow{2}{*}{RF} & \multirow{2}{*}{RENT} & \multirow{2}{*}{GL} & \multicolumn{3}{c}{UBayFS}\\
        & & & & M & F & T\\
        \midrule
        BCW & 6.7 & 3.4 & 10.9 & 6.2 & 2.2 & 4.3 \\
        HD & 6.3 & 3.2 & - & 1.8 & 1.6 & 2.1 \\
        MPE & 9.4 & 24.3 & - & 12.3 & 5.3 & 9.6 \\
        COL & 6.1 & 3.8 & 4.6 & 3.7 & 2.9 & 3.6 \\
        LSVT & 10.0 & 77.9 & 9.0 & 6.4 & 6.7 & 9.6\\
        p53 & 80.2 & 2712.3 & 112.7 & 366.8 & 125.6 & 440.3 \\
        PRO & 29.8 & 1217.2 & - & 370.9 & 232.6 & 708.0 \\
        LEU & 41.5 & 980.9 & - & 263.0 & 160.8 & 549.5 \\
        \bottomrule
    \end{tabular}
\end{table}

\section{Discussion and Conclusion}
\label{sec:discussion}

The presented Bayesian feature selector UBayFS has its strength in combining information from a data-driven ensemble model with expert prior knowledge targeted at the life science domain. The generic framework is flexible in the choice of the elementary feature selector type, allowing a broad scope of applications scenarios by deploying adequate elementary feature selectors, such as those suggested by \cite{Sechidis2018} for semi-supervised or \cite{Elghazel2015} for unsupervised problems. An extension of the presented experiments to multiple classes or multi-label classification problems (one object is not uniquely assigned to one class) is straightforward as well if the elementary feature selector is capable of tackling such datasets, such as \cite{Petkovic2020}.

In general, the choice of the elementary feature selector is a central step when deploying the concept in practice---in particular, performance, stability, and runtime need to be taken into consideration, given the size and structure of a dataset. Still, the main focus of the present work is to discuss the conceptual properties of the framework rather than the individual characteristics of distinct elementary feature selectors. Nevertheless, a broad range of elementary models is used in the presented experiments to provide user guidance in practical setups. The option to build ensembles combining different model types, as discussed by \cite{seijo:ensembleSurvey}, turned out to decrease the stability of UBayFS significantly and is therefore not considered in this study. 

UBayFS presents two ways to account for feature dependencies: a generalized prior model, as well as a decorrelation constraint. The latter effectively restricts the results, such that a simultaneous selection of highly correlated features is penalized. The generalizations of the prior model correct the estimated feature importances by the dependencies---in a low-dimensional scenario, the hyperdirichlet variant is the most accurate choice. However, this variant becomes intractable, if the dimensionality exceeds a few hundred features and requires simulation to determine the expected value in almost any case, preventing from analytically exact solutions. Since our experiments depicted that feature importances obtained from each of the three prior setup types are numerically similar, a conventional Dirichlet setup seems to deliver a sufficiently accurate approximation for high-dimensional datasets. This observation is also supported by the fact that many elementary feature selectors, such as mRMR or HSIC, can account for between-feature correlations, thus reducing the need to consider correlations in the meta-model.
Prior information from experts is introduced via prior feature weights and linking constraints describing between-feature dependencies, represented in a system of side constraints. Via a relaxation parameter, the admissibility is transferred into a soft constraint, which favors solutions that fulfill the constraints, and penalizes violations.
Introducing user knowledge directly into the feature selection process opens new opportunities for data analysis in life science applications. Still, such methodology bears the potential of intentional or unintentional incorrect use: as demonstrated in the experiment, the integration of unreliable or incorrect user knowledge makes the system prone to be steered in a user-defined direction. Users have to be aware that UBayFS may contain subjective inputs to prevent misuse. Thus, precautions must ensure that information provided to the system is sufficiently verified if any critical decisions are based on model output.

Based on the results from extensive experimental evaluations on multiple open-source datasets, a clear benefit of the proposed feature selector lies in the balance between predictive performance and stability. Particularly in life sciences, where few instances are available in high-dimensional datasets, user-guided feature selection can be an opportunity to guide the model to achieve tractable and high-quality results. UBayFS delivers more flexibility to integrate domain knowledge than established state-of-the-art approaches.

A practical limitation of UBayFS is that the runtime is arguably slower than other feature selectors, which becomes an obstacle in very high-dimensional datasets. The use of highly optimized algorithms like the Genetic Algorithm along with an initialization using the suggested Alg. \ref{alg:sampling2} improves this issue. However, it cannot compensate for the computational burden of training multiple elementary models.

\section*{Acknowledgments}

In special we thank Kristian Hovde Liland (NMBU), Cecilia Marie Futsaether (NMBU) and Eirik Malinen (University of Oslo) for their constructive discussions and valuable input for this work, as well as Michael P. Alley (Penn State University) for proof-reading the paper. This work was partly funded by the Norwegian Cancer Society (grant no. 182672-2016).

\begin{appendices}
\section*{Appendix}
\section{Theory}\label{secA1}
\paragraph{Convergence of $ad_{k,\rho}$}
The point-wise convergence $ad_{k,\rho} \underset{\rho\rightarrow\infty}{\longrightarrow} ad_{k}$ holds for arbitrary $\bm{A}\in \mathbb{R}^{K\times N}$ and $\bm{b}\in \mathbb{R}^{K}$ on the domain $\mathcal{D}=\{0,1\}^N$.

\textit{Proof} From the definition of $ad_{k,\rho}(\bm{\delta})$, the claim is trivially fulfilled for $$\bm{\delta}\in\left\{\bm{\delta}'\in\{0,1\}^N:\left(\bm{a}^{(k)}\right)^{T}\bm{\delta}'-b^{(k)}\leq 0\right\}.$$ In the opposite case, we define $\lambda_k$ as $\lambda_k = \left(\bm{a}^{(k)}\right)^{T} \bm{\delta} - b^{(k)} > 0$. It holds that
\begin{align*}
    ad_{k,\rho}(\bm{\delta}) &= \frac{2\xi_{k,\rho}}{1 + \xi_{k,\rho}} \\
    &= \frac{2\exp\left(-\rho \lambda_k\right)}{1 + \exp\left(-\rho\lambda_k\right)}.
\end{align*}
Since $\lambda_k>0$, we obtain $-\rho\lambda_k\underset{\rho\rightarrow\infty}{\longrightarrow} -\infty$, and thus $\xi_{k,\rho} = \exp\left(-\rho\lambda_k\right)\underset{\rho\rightarrow\infty}{\longrightarrow} 0$. It follows that $ad_{k,\rho}(\bm{\delta})\underset{\rho\rightarrow\infty}{\longrightarrow} 0$. Hence, we have shown a point-wise convergence of 
$$ad_{k,\rho}(\bm{\delta}) \underset{\rho\rightarrow\infty}{\longrightarrow} \left\{\begin{array}{cc}1 & \text{if}~ \lambda_k \leq 0 \\
0 & \text{if}~ \lambda_k > 0, \end{array}\right.$$
which equals to $ad_{k}$ on the domain $\mathcal{D}$.

\paragraph{Generalizations of the Dirichlet distribution}
In Section \ref{sec:prior}, we discuss the possibility to replace the Dirichlet distribution with one out of two generalized variants:
\begin{itemize}
    \item the generalized Dirichlet distribution, and
    \item the hyperdirichlet distribution.
\end{itemize}
Both variants preserve the conjugate prior property with respect to the multinomial likelihood, as explained by the according authors. In this part, we provide a short overview on the probability density functions, parameters and (posterior) expected values of these distributions, as these quantities are relevant for the UBayFS setup.

The standard Dirichlet distribution, see e.g. \cite{degroot2005optimal}, is commonly defined by the probability density function
\begin{equation}
    f_{\text{Dir}}(\bm{\theta};\bm{\alpha})=\frac{1}{B(\bm{\alpha})}\prod\limits_{n=1}^{N}\theta_n^{\alpha_n-1},
\end{equation}
where $B(\bm{\alpha})=\frac{\prod\limits_{n=1}^{N}\Gamma(\alpha_n)}{\Gamma\left(\sum\limits_{n=1}^{N}\alpha_n\right)}$ denotes the multivariate beta function. Due to the simple parameter update in the inference step, we obtain the posterior expected value
$$ \mathbb{E}_{\bm{\theta}}\left[\bm{\theta}\right] =  \frac{1}{\Vert \bm{\alpha}^{\circ}\Vert_1}\bm{\alpha}^{\circ},$$
where $\bm{\alpha}^{\circ}=\bm{\alpha}+\bm{\Delta}$.

In essence, the generalized Dirichlet distribution by \cite{wong98} adds an additional parameter vector $\bm{\beta}\in\mathbb{R}^{N-1}$ to the parameter vector $\bm{\alpha}$ from the Dirichlet distribution and is defined via the probability density 
\begin{equation}
    f_{\text{gDir}}(\bm{\theta}')=\prod\limits_{n=1}^{N-1}\frac{1}{B(\alpha_n,\beta_n)}\left(\theta_n'\right)^{\alpha_n-1}\left(1-\sum\limits_{i=1}^{n}\theta_i'\right)^{\gamma_n},
\end{equation}
where $B(\alpha_n,\beta_n)=\frac{\Gamma(\alpha_n)\Gamma(\beta_n)}{\Gamma(\alpha_n+\beta_n)}$, $\gamma_n=\beta_n-\alpha_{n+1}-\beta_{n+1}$ for $n\in[N-2]$, and $\gamma_{N-1} = \beta_{N-1}-1$. In contrast to the standard Dirichlet setting, the distribution is defined on the $N-1$-dimensional space, relaxing the side constraint $\Vert \bm{\theta}\Vert_1=1$ to $\Vert \bm{\theta}'\Vert_1 \leq 1$, $\bm{\theta'}\in\mathbb{R}^{N-1}$ --- both are equivalent, if $\theta_n = \theta_n'$ for $n\in[N-1]$, and $\theta_N = 1-\sum\limits_{n=1}^{N-1}\theta_n'$. The posterior expected value for the generalized Dirichlet distribution is given in closed-form by
$$ \left(\mathbb{E}_{\bm{\theta}}\left[\bm{\theta}\right]\right)_n =  \left\{\begin{array}{ll}
     \frac{\alpha_n + \Delta_n}{\alpha_n + \beta_n + \nu_n} & n=1\\
     \frac{\alpha_n + \Delta_n}{\alpha_n + \beta_n + \nu_n}\prod\limits_{i=1}^{n-1} \frac{\beta_i + n_{i+1}}{\alpha_i + \beta_i + n_i} & n=2,\dots,N-1 \\
     \prod\limits_{i=1}^{N-1} \frac{\beta_i+n_{i+1}}{\alpha_i + \beta_i + \nu_i} & n=N,
\end{array}\right.$$
where $\nu_n=\sum\limits_{i=n}^{N}\Delta_i$, see \cite{wong98}.

An even more general version is the hyperdirichlet distribution by \cite{hankin}, who characterizes the distribution by the probability density function
\begin{equation}
    f_{\text{hDir}}(\bm{\theta})\propto\left(\prod\limits_{n=1}^{N}\theta_n\right)^{-1}\prod\limits_{G\in \mathcal{P}([N])}\left(\sum\limits_{i\in G}\theta_i\right)^{\mathcal{F}(G)},
\end{equation}
where $\mathcal{P}(.)$ denotes the power set and $\mathcal{F}(G)$ denotes the parameter for each possible subset of $[N]$. Since the closed-form expression of the expected value involves the normalization constant, which is intractable in practical high-dimensional setups, we deploy the Metropolis-Hastings (MH) algorithm implemented in \cite{hanking:r} to sample from the hyperdirichlet distribution and determine the expected value empirically from the sample mean.

\section{Experimental datasets}\label{secA2}

All real-world datasets are publicly available (status: 12/2021), see Tab. \ref{tab:supp_datasets}. For datasets with block structure (BCW, COL, LSVT and p53), block indices are given in Tab. \ref{tab:bcw_features}.
\begin{table}[h!]
\caption{Dataset sources.}
    \label{tab:supp_datasets}
\begin{tabular}{ll}
    \toprule
    name & link \\
    \midrule
    HD & \footnotesize{\url{https://archive.ics.uci.edu/ml/datasets/heart+disease}} \\
    BCW &  \footnotesize{\url{https://archive.ics.uci.edu/ml/datasets/breast+cancer+wisconsin+(diagnostic)}}\\
    MPE &  \footnotesize{\url{https://archive.ics.uci.edu/ml/datasets/Mice+Protein+Expression}}\\
    COL & \footnotesize{\url{https://github.com/cran/gglasso}}\\
    LVST & \footnotesize{\url{https://archive.ics.uci.edu/ml/datasets/LSVT+Voice+Rehabilitation}}\\
    p53 & \footnotesize{\url{https://archive.ics.uci.edu/ml/datasets/p53+Mutants}}\\
    LEU & \footnotesize{see R package \textit{spls} \cite{r:spls}}\\
    PRO & \footnotesize{see R package \textit{propOverlap} \cite{r:propOverlap}}\\
    \bottomrule
\end{tabular}
\end{table}

\begin{table}[ht]
    \centering
    \caption{Block indices for datasets with block structure. Feature names indicate the column name patterns, which is used for defining blocks.}
    \label{tab:bcw_features}
    \begin{tabular}{l|l|l|l}
    \toprule
    dataset & block no & indices & feature names \\
    \midrule
    \multirow{3}{*}{BCW} & 1 & 1-10 & mean\\
    & 2 & 11-20 & error\\
    & 3 & 21-30 & worst\\
    \midrule
    \multirow{4}{*}{COL} & 1 & 1-5 & \\
    & 2 & 6-10 & \\
    & \vdots & \vdots & \\
    & 20 & 96-100& \\
    \midrule
    \multirow{14}{*}{LSVT} & 1 & 97-124 & Delta \\
    & 2 & 160-179, 200-219, 251-270, 291-310 & det \\
    & 3 & 129-139, 220-230 & E \\
    & 4 & 140-159, 180-199, 231-250, 271-290 & entropy\\
    & 5 & 62-67 & GNE\\
    & 6 & 52-53 & HNR\\
    & 7 & 77-82 & IMF\\
    & 8 & 1-30 & jitter\\
    & 9 & 84-96 & MFCC \\
    & 10 & 54-55 & NHR \\
    & 11 & 56-58 & OQ\\
    & 12 & 31-51 & shimmer \\
    & 13 & 68-76 & VFER \\
    & 14 & 59-61, 83, 125-128 & other \\
    \midrule
    \multirow{2}{*}{p53} & 1 & 1-4826 & \\
    & 2 & 4827-5408 & \\
    \bottomrule
    \end{tabular}
\end{table}
\end{appendices}

\bibliographystyle{abbrv}
\bibliography{references}

\end{document}